\documentclass[11pt]{article}
\usepackage[final]{acl}
\usepackage{times}
\usepackage{latexsym}
\usepackage{newtxtext}
\usepackage[T1]{fontenc}
\usepackage[utf8]{inputenc}
\usepackage{microtype}
\usepackage{graphicx}
\usepackage{xcolor}
\usepackage[most]{tcolorbox}
\usepackage{booktabs}
\usepackage{tabularx}
\usepackage{multirow}
\usepackage{siunitx}
\usepackage{amsmath,amssymb}
\usepackage{enumitem}
\usepackage{subcaption}
\usepackage{algorithm}
\usepackage{algpseudocode}
\usepackage{pgfplots}
\pgfplotsset{compat=1.18}
\usepackage{fvextra}
\usepackage{silence}
\DefineVerbatimEnvironment{PromptListing}{Verbatim}{
  breaklines=true,
  breakanywhere=true,
  fontsize=\footnotesize
}

\newtcolorbox{promptbox}[1]{
  breakable,
  colback=gray!3,
  colframe=gray!40,
  boxrule=0.4pt,
  arc=2pt,
  left=6pt,
  right=6pt,
  top=6pt,
  bottom=6pt,
  title={#1},
  fonttitle=\bfseries,
}

\usepackage{hyperref}

\title{RAGTurk: Best Practices for Retrieval Augmented Generation in Turkish}

\author{
  Süha Kağan Köse\textsuperscript{1}, Mehmet Can Baytekin\textsuperscript{1}, Burak Aktaş\textsuperscript{1}, Bilge Kaan Görür\textsuperscript{1}, \\
  \textbf{Evren Ayberk Munis}\textsuperscript{2}, \textbf{Deniz Yılmaz}\textsuperscript{3}, \textbf{Muhammed Yusuf Kartal}\textsuperscript{4}, \textbf{Çağrı Toraman}\textsuperscript{3} \\
  \textsuperscript{1}Roketsan Inc., Artificial Intelligence Technologies Unit, Turkey  \\
  \textsuperscript{2}Politecnico di Torino, Italy \\
  \textsuperscript{3}Middle East Technical University, Computer Engineering Department, Turkey \\
  \textsuperscript{4}TOBB University of Economics and Technology, AI Engineering Department, Turkey \\
  \texttt{kagan.kose@roketsan.com.tr}, 
  \texttt{can.baytekin@roketsan.com.tr} \\
  \texttt{burak.aktas@roketsan.com.tr},
  \texttt{kaan.gorur@roketsan.com.tr} \\
  \texttt{evrenayberk.munis@studenti.polito.it},
  \texttt{deniz.yilmaz\_12@metu.edu.tr} \\
  \texttt{m.kartal@etu.edu.tr},
  \texttt{ctoraman@metu.edu.tr}
}

\begin{document}
\maketitle

\begin{abstract}
Retrieval-Augmented Generation (RAG) enhances LLM factuality, yet design guidance remains English-centric, limiting insights for morphologically rich languages like Turkish. We address this by constructing a comprehensive Turkish RAG dataset derived from Turkish Wikipedia and CulturaX, comprising question--answer pairs and relevant passage chunks. We benchmark seven stages of the RAG pipeline—from query transformation and reranking to answer refinement—without task-specific fine-tuning. Our results show that complex methods like HyDE maximize accuracy (85\%) that is considerably higher than the baseline (78.70\%). Also a Pareto-optimal configuration using Cross-encoder Reranking and Context Augmentation achieves comparable performance (84.60\%) with much lower cost. We further demonstrate that over-stacking generative modules can degrade performance by distorting morphological cues, whereas simple query clarification with robust reranking offers an effective solution.\footnote{Links to our datasets and source code are available at: \url{https://github.com/metunlp/ragturk}}
\end{abstract}

\section{Introduction}

Large Language Models (LLMs) perform strongly across many NLP tasks, yet they struggle when queries require current, domain-specific, or verifiable information. Retrieval-Augmented Generation (RAG) mitigates these limitations by incorporating external evidence during generation~\cite{lewis2020rag}. Over time, RAG has developed into a modular pipeline—spanning from query transformation to answer refinement—whose components collectively shape system performance~\cite{gupta2024rag_survey, zhao2024rag_aigc_survey, liu2025ragSurvey}. Many studies have improved individual stages of this pipeline: dense retrievers~\cite{karpukhin2020dpr}, late-interaction models~\cite{khattab2020colbert}, LLM-based query expansion~\cite{gao2023hyde, li2025qeSurvey}, cross-encoder reranking~\cite{nogueira2019bert_rerank}, and hierarchical retrieval~\cite{tao-etal-2025-treerag}. Recent work also emphasizes iterative reasoning in RAG~\cite{asai2024selfrag, jiang-etal-2023-active} and holistic evaluation metrics beyond answer accuracy alone~\cite{yu2024evaluationrag, es-etal-2024-ragas}.

However, nearly all prior RAG research targets English. For morphologically rich and moderately resourced languages like Turkish, the behavior of RAG systems remains largely unknown. Turkish morphology, flexible word order, and variation across sources introduce retrieval and grounding challenges not reflected in English benchmarks. Existing work examines isolated components—retrievers~\cite{bikmaz2025bridging}, cultural QA~\cite{simsek2025turkishculturalrag}, embeddings~\cite{ezerceli2025turkembed4retrieval}, and hallucination detection~\cite{tas2025turklettucedetect}—but not full pipelines evaluated end-to-end on curated Turkish benchmarks. Relevant background on Turkish NLP challenges and benchmark evaluation appears in~\cite{hakkani2002statistical, oflazer2014turkish, umutlu-etal-2025-evaluating}.

Meanwhile, the broader RAG ecosystem is shifting toward pipeline-level optimization. Frameworks such as AutoRAG~\cite{kim2024autorag}, DSPy-RAG~\cite{khattab2024dspy}, GraphRAG~\cite{edge2024graphrag} and RAGSmith~\cite{kartal2025ragsmith} show that performance depends on coordinated component interaction rather than any single module. Yet these systems are also English-centric, leaving open questions about cross-linguistic transferability.

To address this gap, we present the first systematic, end-to-end evaluation of RAG pipeline components for Turkish on a dataset consisting of two parts with similar properties (well-formed Turkish text and grounded question--answer pairs with verifiable evidence passages) but different sources. Using 4{,}891 Turkish Wikipedia articles and 6{,}305 Turkish web articles derived from CulturaX~\citep{nguyen-etal-2024-culturax}, we assess seven core pipeline stages under a unified protocol.

Our contributions are:
\begin{itemize}
    \item \textbf{A comprehensive Turkish RAG benchmark:} we construct a unified dataset sourced from Turkish Wikipedia and CulturaX, generating grounded question--answer pairs with gold evidence passages across factual and interpretive question types.
    \item \textbf{A systematic, end-to-end pipeline study for Turkish:} we benchmark seven core stages of a modern RAG stack (query transformation, candidate re-ranking, filtering \& selection, context augmentation, condensation, prompt composition, and answer refinement) under a unified protocol.
    \item \textbf{Optimized recipes and reproducible release:} we distill actionable strategies—identifying a Pareto-optimal configuration that balances accuracy and efficiency while highlighting the risks of over-stacking LLM modules—and release the datasets, prompts, configuration files, and evaluation scripts to support reproducible Turkish-RAG research.
\end{itemize}

\section{Related Work}
Retrieval-Augmented Generation (RAG) strengthens Large Language Models (LLMs) with grounded, domain-specific information. Since Lewis et al.~\cite{lewis2020rag}, RAG has developed into a modular pipeline where choices across components (from query transformation to retrieval, reranking/selection, context construction/condensation, prompting, and answer refinement) jointly determine performance~\cite{gupta2024rag_survey, zhao2024rag_aigc_survey, liu2025ragSurvey}. Recent evaluation work correspondingly argues for holistic assessment beyond generation quality, emphasizing retrieval effectiveness, grounding, attribution, and factual reliability~\cite{yu2024evaluationrag, es-etal-2024-ragas}.

\paragraph{Retriever Architectures and Query Transformation.}
Dense retrieval replaces sparse lexical matching with neural representations; DPR~\cite{karpukhin2020dpr} demonstrates strong gains over BM25 on knowledge-intensive tasks, and hybrid dense+lexical approaches further improve robustness across query types~\cite{lin2021pyserini}. Complementary work improves recall via query reformulation/expansion: HyDE~\cite{gao2023hyde} generates hypothetical documents to bridge lexical gaps, while surveys of query expansion and multi-query strategies show consistent benefits for ambiguous or underspecified queries~\cite{li2025qeSurvey}.

\paragraph{Candidate Re-ranking, Context Selection, and Condensation.}
Reranking provides finer-grained relevance estimates after retrieval; cross-encoders (often BERT-based) remain the standard for high-precision ranking~\cite{nogueira2019bert_rerank}. Hierarchical/structured approaches such as TreeRAG~\cite{tao-etal-2025-treerag} and related context-selection/condensation methods (e.g., selective extraction, compression, ordering) aim to pack the most useful evidence into limited context windows.

\paragraph{Answer Refinement and Self-Reflective RAG.}
Beyond ``retrieve-then-generate,'' self-reflective methods such as Self-RAG~\cite{asai2024selfrag} and FLARE~\cite{jiang-etal-2023-active} let models check support, trigger additional retrieval, and revise answers, improving factuality and robustness via iterative retrieval--generation.

\paragraph{System-Level Optimization: AutoRAG, DSPy-RAG, and GraphRAG.}
Recent work optimizes RAG end-to-end rather than tuning single components: AutoRAG~\cite{kim2024autorag} automates configuration over retrievers, chunking, query expansion, rerankers, prompts, and post-generation modules; DSPy-RAG~\cite{khattab2024dspy} treats RAG assembly as optimization over declarative modules; and GraphRAG~\cite{edge2024graphrag} uses graph-based indexing and hierarchical retrieval to exploit document structure. These systems highlight strong component interactions, but are evaluated largely on English, leaving open questions for morphologically rich and underrepresented languages.

\paragraph{RAG and Retrieval in Turkish.}
A comprehensive survey of Turkish NLP resources~\cite{coltekin2023turkish} provides essential background on corpora and lexical resources for the language. RAG research for Turkish is emerging: B{\i}kmaz et al.~\cite{bikmaz2025bridging} analyze retrievers and rerankers for Turkish QA, and \c{S}im\c{s}ek~\cite{simsek2025turkishculturalrag} compares RAG against fine-tuning for culturally grounded QA. Turkish retrieval resources such as TurkEmbed4Retrieval~\cite{ezerceli2025turkembed4retrieval} emphasize language-tailored embeddings, while Turk-LettuceDetect~\cite{tas2025turklettucedetect} highlights the need for grounded, verifiable outputs. In addition, recent Turkish LLM benchmarking efforts such as TurkBench~\cite{toraman2026turkbench} include evaluations of retrieval-augmented generation (RAG); however, they remain limited to general-purpose benchmarking and do not target the task-specific retrieval and reasoning challenges considered here.

However, most Turkish studies focus on isolated components rather than full pipelines, and do not systematically test how design choices transfer to Turkish datasets (e.g., Wikipedia articles and broad-coverage web text such as CulturaX) under rich morphology and source-dependent variation.

\paragraph{Positioning our work.}
To address the lack of non-English RAG benchmarks, we present the first systematic, end-to-end study of Turkish RAG pipelines. We construct a two-part dataset sourced from Turkish Wikipedia and CulturaX, and evaluate seven core pipeline stages—from query transformation to answer refinement—under a unified protocol. Our analysis identifies Pareto-optimal configurations that balance accuracy with efficiency, offering actionable ``recipes'' for Turkish retrieval. We release these resources and findings to support reproducible research in morphologically rich languages.

\section{Dataset Construction}
\label{sec:dataset_construction}

To ensure broad coverage of real-world retrieval scenarios, we construct a unified Turkish RAG benchmark comprising two complementary parts that share high standards for text quality and answerability but differ in source characteristics: the \emph{Web Part (CulturaX)} is a diverse collection of Turkish web pages derived from CulturaX~\citep{nguyen-etal-2024-culturax}, filtered to retain contentful text across a wide range of topics (e.g., everyday life, entertainment, news), while the \emph{Wikipedia Part} consists of Turkish Wikipedia articles providing encyclopedic reference text with stronger emphasis on biography, STEM, and history.

\subsection{Corpus Acquisition and Filtering}
\label{sec:web_filtering}
To ensure valid evaluation, we filter raw crawls to keep only contentful, answerable documents. For the Web Part, we start from CulturaX and sample candidate Turkish pages.
We then apply a two-stage filtering procedure guided by LLM-based judgments to ensure high retrieval utility:

\begin{enumerate}[leftmargin=*,itemsep=2pt]
  \item \emph{URL-only filtering (triage).}
  Given only the base website URL, an LLM estimates whether the site likely contains \emph{valuable, informational} content---operationally, pages with factual statements that could answer user queries (\emph{valuable}) and substantive prose rather than navigation menus, link lists, or boilerplate (\emph{informational}). This stage acts as a low-cost pre-filter to exclude low-value page types such as pure landing pages, navigation hubs, or aggregator farms.

  \item \emph{Content-based filtering (page-level quality).}
  For pages passing stage (1), we fetch the page text and apply a second LLM filter that checks \emph{coherence}, \emph{content depth}, and \emph{utility}. We retain pages that are understandable and content-rich (e.g., blog posts, forum discussions, news articles) with \emph{good quality}. We reject pages that are \emph{spam} (keyword stuffing, bots), \emph{thin} (insufficient content), or \emph{boilerplate} (largely non-textual content).
\end{enumerate}

\paragraph{Prompt design for filtering.}
Filtering prompts are designed to retain answerable passages while excluding content that trivially breaks evaluation. We use gemini~2.5~flash~\citep{google2025gemini25models} for both stages; full prompts are in Appendix~\ref{app:prompts} (see Prompt~\ref{prompt:url_filter} and Prompt~\ref{prompt:content_filter}).

\paragraph{Final Web corpus.}
After filtering, we retain 6{,}305 web pages. We convert each page to Markdown by preserving the main textual sections and mapping prominent HTML headings to Markdown headers.

\paragraph{Website Frequency Analysis.}
\label{sec:site_freq}
We report the base-domain frequency to ensure the Web Part is not dominated by a single source. The top domains include popular sites like sikayetvar.com and haberler.com, but the top 10 domains cover only $\sim$19.6\% of documents, ensuring diversity (see Appendix~\ref{app:full_results}, Table~\ref{tab:top_domains_app} for full list).

\paragraph{Wikipedia Articles.}
For the Wikipedia articles, acquisition is straightforward given the structured nature of the source.
We randomly sample Turkish Wikipedia pages, exclude short articles (< 300 chars), and retrieve plain-text sections.
After filtering, we retain 4{,}891 articles and convert them to Markdown, mapping sections to headers.

\subsection{Header-Aware Chunking}
\label{sec:chunking}

We apply a header-aware chunking strategy. Each chunk inherits document and section context (e.g., page title and section path).
Long segments are split if they exceed 1{,}000 characters. We tried different thresholds and found this character limit provides the best balance between (i) preserving enough local context for answerability and (ii) limiting topic drift.
This tokenizer-agnostic limit is particularly stable for Turkish, where agglutination packs more information into fewer whitespace-delimited tokens.

\subsection{Topic Categorization}
\label{sec:categorization}

We annotate each document with a unified topic label using an LLM-based classifier using again gemini~2.5~flash~\citep{google2025gemini25models} as the LLM provider. We adapt our categorization approach and prompt design considerations from prior work on large-scale dataset construction~\citep{gao2020pile,soldaini-etal-2024-dolma,weber2024redpajama,penedo2024fineweb,wenzek-etal-2020-ccnet,elazar2023wimbd}. Full prompts are in Appendix~\ref{app:prompts} (see Prompt~\ref{prompt:topic_classifier}).

\paragraph{Topic taxonomy.}
We define 10 broad topic categories (Table~\ref{tab:topic_stats}).
The distribution highlights the complementary nature of the dataset:
The \emph{Web Part} leans towards \textit{Everyday Life}, \textit{Entertainment}, and \textit{Politics}, reflecting the conversational web.
The \emph{Wikipedia Part} has higher coverage of \textit{Biography}, \textit{STEM}, and \textit{History}.
Taken together, the dataset spans the full range of user queries, from checking facts on public figures to navigating forum advices.

\begin{table}[t]
  \centering
  \caption{Topic statistics per dataset part and overall totals. Web/Wikipedia percentages are computed w.r.t. the full dataset ($N{=}11{,}196$).}
  \label{tab:topic_stats}
  \small
  \setlength{\tabcolsep}{6pt}
  \begin{tabularx}{\linewidth}{>{\raggedright\arraybackslash}X r|r|rr}
  \toprule
  \multirow{2}{*}{\textbf{Topic}} &
  \multicolumn{1}{c|}{\textbf{Web}} &
  \multicolumn{1}{c|}{\textbf{Wikipedia}} &
  \multicolumn{2}{c}{\textbf{Total}} \\
  \cmidrule(lr){2-2}\cmidrule(lr){3-3}\cmidrule(lr){4-5}
  & \textbf{\%} & \textbf{\%} & \textbf{\#} & \textbf{\%} \\
  \midrule
  Entertainment & 10.7 & 8.1  & 2{,}104 & 18.8 \\
  Biography     & 1.4  & 13.9 & 1{,}712 & 15.3 \\
  Everyday Life & 12.4 & 0.2  & 1{,}414 & 12.6 \\
  STEM          & 5.2  & 6.6  & 1{,}322 & 11.8 \\
  Politics      & 9.9  & 1.6  & 1{,}298 & 11.6 \\
  Professional  & 7.4  & 1.0  & 939     & 8.4 \\
  History       & 3.8  & 4.5  & 925     & 8.3 \\
  Organizations & 3.4  & 2.2  & 627     & 5.6 \\
  Geography     & 1.0  & 3.7  & 525     & 4.7 \\
  Humanities    & 0.9  & 1.9  & 309     & 2.8 \\
  Uncategorized & 0.1  & 0.0  & 21      & 0.2 \\
  \midrule
  \textbf{Total} & \textbf{56.3} & \textbf{43.7} & \textbf{11{,}196} & \textbf{100.0} \\
  \bottomrule
  \end{tabularx}
\end{table}

\subsection{Question--Answer Pair Generation}
\label{sec:qa_gen}

We use gpt-oss:120B~\citep{openai2025gptoss} for generation and validate with gemini-2.5-flash~\citep{google2025gemini25models} as an auxiliary consistency check. This cross-model verification reduces obvious hallucinations and off-topic generations, though it does not guarantee perfect grounding. We adapt our QA generation approach and prompt design considerations from prior work on large-scale dataset construction~\citep{rajpurkar-etal-2016-squad,yang-etal-2018-hotpotqa,bloom1956taxonomy,anderson2001taxonomy,zheng2023judging,liu-etal-2023-g-eval}. Full prompts are in Appendix~\ref{app:prompts} (see Prompt~\ref{prompt:qa_generation}).
Table~\ref{tab:corpus_stats} reports the final statistics.

\begin{table}[t]
\centering
\caption{Corpus and QA statistics for the complementary parts.}
\label{tab:corpus_stats}
\small
\setlength{\tabcolsep}{4pt}
\begin{tabularx}{\linewidth}{%
  >{\raggedright\arraybackslash}X
  >{\hsize=0.85\hsize\raggedleft\arraybackslash}X
  >{\hsize=1.15\hsize\raggedleft\arraybackslash}X
  >{\hsize=1.00\hsize\raggedleft\arraybackslash}X}
\toprule
\textbf{Statistic} & \textbf{\mbox{Web Part}} & \textbf{\mbox{Wikipedia Part}} & \textbf{\mbox{Total}}\\
\midrule
Articles                & 6{,}305 & 4{,}891 & 11{,}196\\
Characters              & 9{,}933{,}523  & 22{,}821{,}895 & 32{,}755{,}418\\
Char./Article      & 1{,}575.50     & 4{,}666.10 & 2{,}925.46\\
\midrule
Chunks                  & 15{,}985       & 42{,}304 & 58{,}289\\
Chunks/Article          & 2.54           & 8.65 & 5.21\\
Char./Chunk        & 695.61         & 598.81 & 561.97\\
\midrule
Questions               & 10{,}682       & 9{,}777 & 20{,}459\\
Questions/Article       & 1.69           & 2.00 & 1.83\\
Factual                 & 6{,}522        & 5{,}196 & 11{,}718\\
Interpretation          & 4{,}160        & 4{,}581 & 8{,}741\\
\bottomrule
\end{tabularx}
\end{table}

\paragraph{Interpreting corpus and QA statistics.}
The statistics in Table~\ref{tab:corpus_stats} confirm that the two parts offer complementary structural challenges.
The \emph{Web Part} contains more documents but with shorter average length, emphasizing precision in a broad search space.
The \emph{Wikipedia Part} contains fewer but much longer documents, requiring effective passage retrieval within dense, sectioned text.
By covering both, and maintaining a balanced mix of \textit{Factual} and \textit{Interpretation} questions, the benchmark provides a robust testbed for Turkish RAG systems across the spectrum of quality Turkish text.

\section{Methodology and Optimization}
\label{sec:methodology}

Optimizing RAG pipelines requires navigating a vast combinatorial space of design choices. To address this, we adopt a two-step approach: first, we define a comprehensive design space of candidate methods (Section~\ref{sec:rag_space}); second, we employ a budgeted genetic search (Section~\ref{sec:ga}) to efficiently identify high-performing configurations without exhaustive enumeration.

\subsection{RAG Design Space}
\label{sec:rag_space}

\begin{figure}[h!]
    \centering
    \includegraphics[width=\columnwidth]{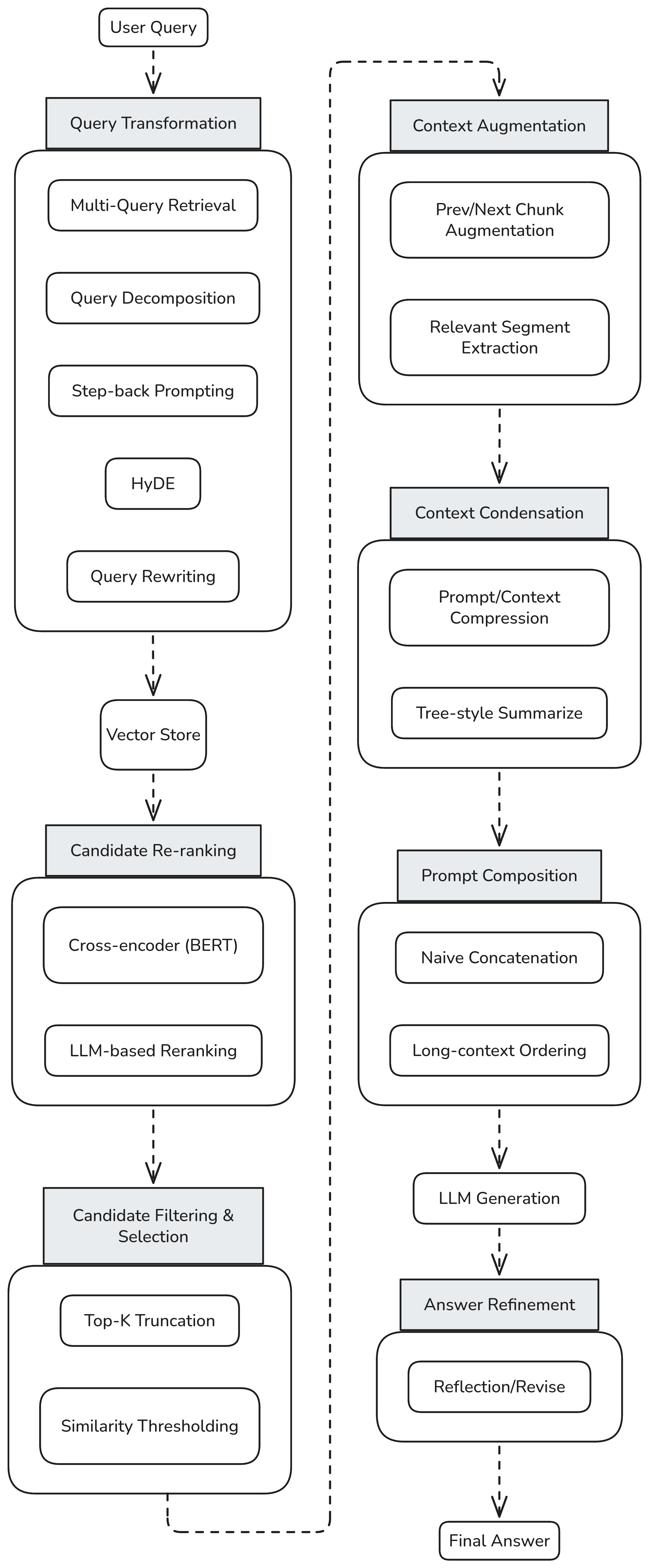}
    \caption{Overview of the RAG design space.}
    \label{fig:rag_pipeline}
\end{figure}

We evaluate a modular RAG pipeline (Figure~\ref{fig:rag_pipeline}) and vary methods within seven technique families while holding constant the rest of the system (chunking policy, index configuration, and prompt structure) to isolate which design choices drive performance.

\paragraph{Selection rationale.}
Because the RAG literature is rapidly expanding, we focus on a representative set of techniques that cover the most common control points in end-to-end RAG (query-side, retrieval/reranking, context selection/compression, and generation-time refinement), are widely used or frequently reported as effective, and can be implemented reproducibly as modular components under a consistent evaluation protocol. A consistent protocol is vital for emerging ecosystems like Turkish~\citep{bikmaz2025bridging,simsek2025turkishculturalrag,ezerceli2025turkembed4retrieval,tas2025turklettucedetect,hakkani2002statistical,ezerceli2025turkembed4retrieval,umutlu-etal-2025-evaluating}, where rigorous standards are often lacking~\citep{umutlu-etal-2025-evaluating}. For techniques that require prompts (e.g., rewriting/decomposition, HyDE-style generation, reflection/revision, LLM reranking), we use the default prompts and hyperparameters recommended in the corresponding papers.

\paragraph{Query transformation.}
\emph{Multi-Query Retrieval:} Generate multiple semantically diverse rewrites of the user query and retrieve for each, then merge results to improve recall~\citep{rackauckas2024ragfusion}.
\emph{Query Decomposition:} Break a complex question into simpler sub-questions, retrieve for each part, and aggregate evidence before answering~\citep{zheng2024stepback}.
\emph{Step-back Prompting:} Ask a more general ``step-back'' question to retrieve high-level background context that helps answer the original query~\citep{zheng2024stepback}.
\emph{HyDE:} Synthesize a hypothetical answer document from the query and use it as a retrieval query to better match relevant passages~\citep{gao2023hyde}.
\emph{Query Rewriting / LLM-based expansion:} Rewrite and/or expand the query with additional keywords and paraphrases to reduce lexical mismatch and improve retrieval~\citep{wang2023query2doc,mao2021gar}.

\paragraph{Candidate re-ranking.}
\emph{Cross-encoder reranking (BERT):} Score each (query, passage) pair jointly with a cross-encoder and reorder retrieved passages by predicted relevance~\citep{nogueira2019bert_rerank}.
\emph{LLM-based reranking:} Use an instruction-tuned LLM to judge relevance of candidates and reorder/retain the most useful passages for answering~\citep{lewis2020rag}.

\paragraph{Candidate filtering \& selection.}
\emph{Top-$K$ truncation:} Keep only the $K$ highest-ranked retrieved passages to control context budget and reduce noise~\citep{karpukhin2020dpr,lewis2020rag}.
\emph{Similarity thresholding:} Discard candidates below a similarity cutoff to avoid injecting weakly related context into generation~\citep{karpukhin2020dpr,lewis2020rag}.

\paragraph{Context augmentation.}
\emph{Prev/next chunk augmentation (multi-granular context):} Expand a retrieved chunk with its neighbors or multi-scale spans to recover lost local coherence from segmentation~\citep{liu2025lgmgc}.
\emph{Relevant segment extraction:} Extract only the most relevant spans within retrieved documents to maximize signal per token in the context window~\citep{liu2025lgmgc}.

\paragraph{Context condensation.}
\emph{Prompt/context compression \& redundancy pruning:} Compress passages and remove redundant content so more unique evidence fits within the model’s context length~\citep{jiang2023llmlingua,li2023selectivecontext}.
\emph{Tree-style summarize / iterative refine:} Summarize evidence hierarchically or refine a running summary over multiple steps to preserve key facts under tight budgets~\citep{madaan2023selfrefine}.

\paragraph{Prompt composition.}
\emph{Naive concatenation (RAG baseline):} Concatenate the selected passages into a single context block and prompt the model to answer grounded in that context~\citep{lewis2020rag}.
\emph{Long-context ordering (``lost in the middle''):} Vary the ordering/placement of evidence in long prompts to study position effects where mid-context facts are under-attended~\citep{liu2024lostmiddle}.

\paragraph{Answer refinement.}
\emph{Reflection/Revise:} Generate an initial answer, critique it against the retrieved evidence, and revise to fix omissions or inconsistencies~\citep{shinn2023reflexion,madaan2023selfrefine}.

\subsection{Genetic Algorithm to Determine Effective Combinations}
\label{sec:ga}

The modular RAG design space is highly combinatorial, yielding $\sim$1,296 possible pipelines in our setting---too many for exhaustive evaluation. To efficiently navigate this space, we employ a constrained genetic algorithm (GA) inspired by \citet{kartal2025ragsmith}. GAs are well-suited for combinatorial optimization \citep{holland1975adaptation,goldberg1989genetic}, allowing us to identify near-optimal solutions by evaluating only $\sim$200 pipelines (approx. 15\% of the search space). We supplement this automated search with manual evaluation of established baselines. The GA evolves a population of pipelines by (i) evaluating them on a small query set, (ii) selecting top performers, and (iii) generating new candidates via crossover and mutation, enforcing compatibility constraints.

\paragraph{Genome encoding.}
\label{sec:encoding}

We encode a RAG pipeline as a discrete genome:
\[
g = (m_1, m_2, \ldots, m_F)
\]
where $F$ is the number of technique families. Each gene $m_f \in \mathcal{M}_f$ selects exactly one method from family $f$ (optionally including a \texttt{None} choice to disable that slot).

\paragraph{Fitness function.}
\label{sec:fitness}

We optimize a composite objective that balances retrieval effectiveness and answer quality on an evaluation subset $\mathcal{Q}$. Let $R(g)$ denote a retrieval metric (e.g., nDCG@$k$ or Recall@$k$) and $G(g)$ a generation metric (e.g., judged faithfulness/accuracy). We compute both metrics on $\mathcal{Q}$ and normalize them to comparable scales (denoted by $\widetilde{\cdot}$). The fitness of genome $g$ is:
\begin{equation}
\label{eq:fitness}
\mathrm{Fit}(g) = \alpha \cdot \widetilde{R}(g) + (1-\alpha)\cdot \widetilde{G}(g).
\end{equation}
We set $\alpha=0.5$ to weight retrieval and generation equally, reflecting that strong RAG performance requires both (i) retrieving relevant evidence and (ii) producing a faithful and accurate response conditioned on that evidence. This equal weighting also avoids over-specializing the search toward pipelines that optimize retrieval quality at the expense of answer quality (or vice versa).

\paragraph{GA Procedure and evaluation budget.}
\label{sec:ga_budget}

We run the GA for a small number of generations to identify strong pipelines under a fixed evaluation budget. Concretely, we use population size $P=20$ and $G=10$ generations. Each candidate genome is evaluated on a randomly sampled set of $|\mathcal{Q}|=100$ questions per domain. This yields a total of $P \times G$ candidate evaluations while keeping per-candidate evaluation lightweight. This GA evaluation serves as the primary empirical basis for the paper's best-practice conclusions (Section~\ref{sec:best_practices}); the full algorithm is detailed in Appendix~\ref{app:full_results} (Algorithm~\ref{alg:ga_app}).

\paragraph{Reproducibility.}
All parameter values used in the genetic search (including selection and elitism settings, crossover/mutation operators and rates, constraint checks, and random seeds) as well as the full Python implementation of the algorithm used in our experiments are shared in the code repository.

\subsection{Experimental Setup and Metrics}
\label{sec:setup}

We select metrics to (i) capture complementary failure modes, (ii) align with common practice to ease comparison across RAG systems, and (iii) reduce sensitivity to any single noisy automatic signal. Concretely, for retrieval we report a mix of \emph{coverage}-oriented metrics (Recall@5) and \emph{ranking-quality} metrics (mAP, nDCG@5, MRR), since downstream generation depends both on whether evidence is retrieved at all and on how highly it is ranked. For generation, we combine an embedding-based semantic similarity signal with an LLM-as-a-judge score to balance paraphrase-tolerant matching with a more holistic assessment of correctness and answer quality; this choice follows the recommendation to use multiple complementary evaluation criteria, including LLM-judge style assessments, when analyzing benchmark and system outputs~\citep{umutlu-etal-2025-evaluating}.

\subsubsection{Retrieval Metrics}
\label{sec:retrieval_metrics}

We evaluate retrieval against the ground-truth evidence passages used to generate each Q\&A.
For a query $q$ with relevant passages $\mathcal{D}^\star$ and the ranked list $\pi_k(q)$ of top-$k$ retrieved passages:
We compute an overall \emph{retrieval score} as an equally weighted aggregate of Recall@5, mAP, nDCG@5, and MRR (we report the component metrics alongside the aggregate).
\begin{equation}
\label{eq:retrieval_score}
\begin{aligned}
\mathrm{Retrieval}(q) &= \alpha\Big(\mathrm{Recall@}5(q) + \mathrm{mAP}(q) \\
& +\, \mathrm{nDCG@}5(q) + \mathrm{MRR}(q)\Big)
\end{aligned}
\end{equation}
where we set $\alpha=0.25$ to equally weight the four complementary retrieval metrics (coverage and ranking quality) without privileging any single component, consistent with our equal-weight aggregation choices elsewhere.

\subsubsection{Generation Metrics}
\label{sec:gen_metrics}

We measure end-to-end answer quality with two complementary signals and compute an overall \emph{generation score} as their equally weighted aggregate (we also report each component):
(i) \emph{Semantic similarity} (embedding-based similarity between the model answer and the reference answer), and
(ii) \emph{LLM-Judge} (an LLM-based judgment score for answer quality/correctness).
We aggregate them as
\begin{equation}
\label{eq:generation_score}
\mathrm{Generation}(q) = \alpha\,\mathrm{Sim}(q) + (1-\alpha)\,\mathrm{Judge}(q).
\end{equation}
In our experiments we set $\alpha=0.5$ to give equal weight to semantic similarity and the judge signal, reflecting a conservative choice that avoids over-optimizing to either an embedding proxy or a single LLM-based evaluator (analogous to the equal-weight aggregation used in our GA fitness objective).

\subsubsection{Latency and Practicality}
\label{sec:latency}

We report total token usage for each configuration but do not use this metric when optimizing with the GA; we report it to take into account the practicality of the configuration when suggesting best practices.
Table~\ref{tab:compute} records model versions and system settings.

\begin{table}[t]
\centering
\caption{Compute and implementation details.}
\label{tab:compute}
{\small
\setlength{\tabcolsep}{4pt}
\begin{tabularx}{\linewidth}{>{\raggedright\arraybackslash}p{0.36\linewidth} >{\raggedright\arraybackslash}X}
\toprule
\textbf{Component} & \textbf{Setting}\\
\midrule
Embedding model & \href{https://huggingface.co/models?search=embeddinggemma}{embeddinggemma}\\
Generator model & \href{https://huggingface.co/models?search=gpt-oss\%20120B}{gpt-oss:120B}\\
Evaluator model & \href{https://huggingface.co/models?search=gemini-2.5-flash}{gemini-2.5-flash}\\
Reranker & \href{https://huggingface.co/cross-encoder/ms-marco-MiniLM-L-12-v2}{ms-marco-MiniLM-L-12-v2}\\
Hardware & M3 Ultra 80-core GPU\\
\bottomrule
\end{tabularx}
}
\end{table}

\section{Best Practices}
\label{sec:best_practices}
\subsection{Evaluation Protocol}
\label{sec:protocol}

Our evaluation is performed within the GA procedure (Section~\ref{sec:ga}). We sample a stratified random subset of $n=100$ questions (balanced across the Web and Wikipedia parts) from the unified benchmark and use this set both to score candidate genomes during the search and to report the final performance of the best GA-selected configurations. The full benchmark contains $N \approx 20{,}459$ grounded QA pairs; subsampling keeps evaluation tractable while enabling controlled comparison across configurations. This evaluation procedure constitutes the empirical basis for our results.

\paragraph{Why $n=100$ is sufficient.}
In pilot runs, we varied the subset size and observed that performance estimates saturated around $n=100$. Beyond this point, variance reduction was marginal compared to the linear increase in evaluation cost. Thus, $n=100$ serves as an efficient saturation point that yields stable mean estimates and consistent relative rankings of candidate configurations, while keeping the per-candidate evaluation cost low enough for the GA to explore many pipelines under a fixed budget.

\subsection{Overall Performance}
\label{sec:overall}

Table~\ref{tab:main_results} summarizes the performance of the top configurations identified by the genetic search on the unified Turkish RAG benchmark ($n=100$). We compare the baseline against three distinct optimal points found by the GA: a maximum-accuracy configuration, a Pareto-optimal ``best value'' configuration, and a production-friendly configuration. The complete performance results for all noteworthy configurations are provided in Appendix~\ref{app:full_results} (Table~\ref{tab:full_results}).

\begin{table*}[t]
\centering
\caption{Performance of GA-selected configurations on the unified benchmark ($n=100$). We report Overall Score, Retrieval Score, Generation Score, and estimated Token usage per query. The ``High Accuracy'' model achieves the best scores but at high cost, while the ``Pareto Optimal'' model offers the best balance.}
\label{tab:main_results}
{\scriptsize
\setlength{\tabcolsep}{4pt}
\renewcommand{\arraystretch}{0.95}
\begin{tabularx}{\textwidth}{>{\raggedright\arraybackslash}X *{4}{>{\centering\arraybackslash}p{0.12\textwidth}}}
\toprule
\textbf{Configuration} & \shortstack{\textbf{Overall}\\\textbf{Score}} & \shortstack{\textbf{Retrieval}\\\textbf{Score}} & \shortstack{\textbf{Generation}\\\textbf{Score}} & \shortstack{\textbf{Tokens}\\\textbf{(est.)}}\\
\midrule
\textbf{High Accuracy} (HyDE + Cross-encoder Reranking + Tree-style Summarize + Long-context Ordering) & 85.00\% & 0.876 & 0.823 & $\sim$3664 \\
\textbf{Pareto Optimal} (Cross-encoder Reranking + Previous/Next Chunk Augmentation + Long-context Ordering) & 84.60\% & 0.870 & 0.823 & $\sim$1987 \\
\textbf{Production-Friendly} (Query Clarification + Cross-encoder Reranking + Previous/Next Chunk Augmentation) & 80.20\% & 0.901 & 0.704 & $\sim$1738 \\
\textbf{Baseline} (Dense Retrieval + Similarity Thresholding + Naive Concatenation) & 78.70\% & 0.872 & 0.702 & $\sim$1000 \\
\bottomrule
\end{tabularx}
}
\end{table*}

\subsection{Component Analysis and Inferences}
\label{sec:analysis}

Our results highlight several key trade-offs in the design space of Turkish RAG systems.

\paragraph{Maximizing Accuracy.}
If the goal is to maximize the end-to-end score, the winning recipe is a combination of strong query expansion (HyDE), strong reranking (Cross-encoder Reranking), aggressive context compression (Tree-style Summarize), and Long-context Ordering. This configuration achieves the top score of 85.00\%, driven by high generation quality (0.823). However, this comes at a significant cost: approximately $3.7\times$ the token usage of the baseline. HyDE and summarization are computationally expensive, making this approach suitable only when accuracy is paramount and resources are unconstrained.

\paragraph{The Pareto Winner.}
The ``Best Value'' configuration (Cross-encoder Reranking + Previous/Next Chunk Augmentation + Long-context Ordering) achieves 84.60\% overall score, which is only 0.4 points behind the top performer, but with significantly lower overhead. Compared to the baseline, it yields a +5.9 point improvement for $\sim2\times$ tokens. This represents the cleanest recommendation for practical applications: prioritize Cross-encoder Reranking and local context enrichment (Previous/Next Chunk Augmentation) before adopting expensive methods like HyDE.

\paragraph{Lightweight Best Option (Production-Friendly).}
The ``Production-Friendly'' configuration (Query Clarification + Cross-encoder Reranking + Previous/Next Chunk Augmentation) achieves 80.20\% overall score, with Retrieval 0.901 and Generation 0.704, consuming $\sim$1,738 tokens ($\approx$ 1.74$\times$ baseline tokens). 
A short clarification / rewrite step plus cross-encoder reranking and adjacent-context augmentation delivers a strong accuracy gain at modest cost. This likely helps Turkish by reducing morphology-driven ambiguity and improving matching for entity/surface-form variation, while avoiding extra overhead from context reordering.

\paragraph{HyDE and LLM Reranking Costs.}
We find that HyDE is not a ``free lunch'' on Turkish datasets. While it helps bridge semantic gaps, it often balloons costs and can underperform if it introduces noise. For instance, HyDE combined with Tree-style Summarize led to high computational overhead. Similarly, LLM-based Reranking was outperformed by Cross-encoder Reranking in terms of score-per-cost. Cross-encoder Reranking proved to be a strong, reliable default, whereas LLM-based Reranking should be reserved for niche reasoning-heavy cases.

\paragraph{Risks of Over-Stacking.}
Stacking multiple LLM-based modules often degrades efficiency without guaranteeing better performance. A maximalist pipeline combining six complex modules achieved only 79.60\% accuracy, underperforming the simpler ``High Accuracy'' configuration (85.00\%) while consuming heavy token usage. Excessive LLM post-processing in Turkish may distort morphological cues or accumulate errors. From these results, we can recommend adding at most one LLM-heavy stage (e.g., summarization or reflection) only when necessary, avoiding ``stacking everything'' without proven gain.

\paragraph{Over-Filtering Problems.}
Strict filtering combined with segment extraction can severely harm retrieval in Turkish. A pipeline using strict thresholds and segment extraction dropped to 78.30\% overall score (Retrieval 0.755), well below baseline. Morphology and paraphrase variance in noisy web data make hard thresholds risky, causing the system to miss evidence. We recommend preferring reranking over strict filtering unless thresholds are carefully calibrated per domain.

\subsection{Recommendations}
\label{sec:recommendations}

Based on these findings, we propose the following best practices for Turkish RAG:

\begin{itemize}
    \item \emph{Recommended Default:} Use \emph{Cross-encoder Reranking + Previous/Next Chunk Augmentation + Long-context Ordering}. This pipeline offers a strong balance of accuracy and efficiency and should serve as the standard baseline for Turkish RAG experiments.
    
    \item \emph{High-Accuracy:} For leaderboards or applications where score is critical, use \emph{HyDE + Cross-encoder Reranking + Tree-style Summarize + Long-context Ordering}. This requires a higher budget for latency and tokens.
    
    \item \emph{Production-Friendly:} For latency-constrained applications, use \emph{Query Clarification + Cross-encoder Reranking + Previous/Next Chunk Augmentation}. This provides meaningful improvements over naive RAG while maintaining fast response times and low token usage.
\end{itemize}

\noindent These recommendations are specific to Turkish and were derived under our experimental setup. While other morphologically rich languages such as Finnish, Hungarian, and Korean face similar challenges (e.g., morphological richness, agglutination, surface-form variation)~\citep{tsarfaty-etal-2020-spmrl,gerz-etal-2018-relation}, we do not claim that our findings transfer directly; validating generalizability to other languages requires dedicated experiments.

\section{Conclusion and Future Work}
\label{sec:conclusion_future}

We presented an end-to-end, domain-aware study of Turkish RAG across informal web text and Turkish Wikipedia. By benchmarking modular choices across the RAG pipeline, we distill practical, domain-specific configuration guidance and provide resources to support reproducible Turkish-RAG research.

\paragraph{Future Work.}We plan to: (i) explore hybrid RAG within a family instead of using single method from each, (ii) incorporate graph structure for retrieval and query expansion (entity graphs, hyperlink graphs), (iii) scale to larger and more diverse Turkish corpora (news, technical documentation, legal text), (iv) study Turkish morphology-aware retrieval features (e.g., lemma-aware sparse retrieval, morphological analyzers), (v) incorporate more noisy, real-world-like data and unanswerable questions to better assess system robustness (moving beyond our current evaluation on relatively clean questions), and (vi) examine document-side, index-time methods---including pre-embeddings and related pre-computation/caching techniques. We also plan controlled technique-family ablations per domain to quantify marginal gains and interactions (retrieval vs.\ generation) under a fixed evaluation and budget.
Additionally, we plan to conduct a structured error analysis of end-to-end outputs, focusing on Turkish-specific failure modes such as inflection-driven mismatch, over-normalization of informal language, entity drift, and missing evidence in multi-passage questions. A systematic taxonomy and annotated error set will help separate retrieval versus generation errors and guide targeted improvements.

\section{Limitations and Ethical Considerations}
\label{sec:limitations}

\paragraph{Limitations.}
All recommendations in this paper reflect our specific experimental setup (models, prompts,
tokenization, corpus preprocessing, hardware, and context limits). In practice, the best-performing settings can
shift across Turkish corpora due to differences in domain, document length distribution, content quality,
noise/boilerplate, and latency constraints. We therefore position our findings as \emph{best practices for generic
Turkish text retrieval}, and encourage practitioners to re-run a small sweep on their own data to identify the
best point on the quality--latency tradeoff.
While the dataset reflects realistic web content, it is cleaner than typical production RAG pipelines and does not cover specialized domains such as legal or technical documentation; extending the benchmark to such domains is left for future work.
As a defense-industry organization, we cannot release or fully describe some proprietary data sources used during development (e.g., internal enterprise documents); therefore, this paper and the released resources focus on openly available data.

\paragraph{Reproducibility and releases.}
\label{sec:repro}

We release the evaluation datasets, the full QA set with evidence spans, all RAG configuration files, and scripts to reproduce metrics in the code repository.
All recommendations in this paper reflect our specific experimental setup (models, prompts,
tokenization, corpus preprocessing, hardware, and context limits). In practice, the best-performing settings can
shift across Turkish corpora due to differences in domain (formal vs.\ informal), document length distribution,
noise/boilerplate, and latency constraints. We therefore position our findings as \emph{best practices for generic
Turkish text retrieval}, and encourage practitioners to re-run on their own data to identify the
best point on the quality--latency tradeoff.

\paragraph{Ethics.}
Informal web data can contain sensitive or personal content; we recommend careful filtering, redaction, and license-aware release. Because the data are collected from the public internet, they may reflect societal biases and other problematic content; any such content is included for research purposes only and does not reflect the authors' opinions or endorsements. LLM-based filtering can itself introduce bias; we therefore document filtering criteria and provide audit samples where feasible.

\paragraph{Data Provenance and Copyright.}
The Web Part of our dataset is derived from CulturaX~\citep{nguyen-etal-2024-culturax}, a publicly available multilingual corpus; we do not perform independent scraping of websites. Source domains (e.g., haberler.com, sikayetvar.com) were included in CulturaX due to their topical diversity and public accessibility. We release only derived annotations---question--answer pairs, topic labels, and chunk boundaries---rather than redistributing full original articles. This approach aligns with standard research practices for web-derived corpora and respects the original data providers.

\paragraph{Use of Generative AI.}
Generative AI was used solely to assist with language editing. All scientific contributions, data construction, analysis, and interpretations presented in this work are original and were conducted entirely by the authors.

\section*{Acknowledgments}
We gratefully acknowledge support from Roketsan Inc. and the Google Gemini Academic Reward Program, which helped enable the experiments and computing resources used in this study.

\nocite{Aho:72}

\bibliographystyle{acl_natbib}
\bibliography{main}

\onecolumn
\appendix
\section{Prompts and Validation Rubrics}
\label{app:prompts}

\subsection{URL-only Filtering Prompt}
\begin{promptbox}{URL Filtering}
\begin{PromptListing}
Analyze the following website (base URL) and determine its eligibility:

Website: {url}

Evaluate:
1. Does this website likely contain valuable information (educational, informative, useful content)?
2. Is the content on this website likely written in proper language (casual, conversational)?

Based on your analysis of the website domain and typical content, provide:
- Status: "ELIGIBLE" if BOTH conditions are true, otherwise "NOT ELIGIBLE"
- Reason: Brief explanation (1-2 sentences)

Response format:
Status: [ELIGIBLE/NOT ELIGIBLE]
Reason: [Your explanation]
\end{PromptListing}
\label{prompt:url_filter}
\end{promptbox}

\subsection{Content Filtering Prompt}
\begin{promptbox}{Content Filtering}
\begin{PromptListing}
You are a data quality and style evaluator. You will be given TURKISH text taken from a web page, along with the URL it came from.

TASK 1 — EVALUATION
Evaluate whether the text is:
- suitable for a RAG system,
- understandable,
- and "CLEAN" (everyday language; not nonsense/trash).

Definitions:

1) Informality level:
- "clean": Everyday language (blog/forum/social media), but:
  * understandable
  * slightly relaxed yet still structured
  * similar to news-site tone
  * sentences mostly well-formed
  * no heavy slang, no spam
- "nonsense_or_spam": incoherent, random words, bot/spam, only links/hashtags, etc.

2) Quality:
- "good": clear, coherent, not full of spelling errors, topic is followable, usable for RAG
- "bad": too short, messy, major spelling/spam issues, topic not followable

3) ELIGIBLE criteria:
- mostly Turkish
- "Clean"
- Quality must be "good"
- definitely NOT "formal"
- definitely NOT "nonsense_or_spam"
- text length > 100 characters
- mostly about a single topic/theme

IMPORTANT: First do the evaluation and determine Status.

TASK 2 — MARKDOWN CONVERSION (ONLY IF ELIGIBLE)
WARNING: Do this step ONLY if Status: ELIGIBLE. If NOT ELIGIBLE, do NOT convert to markdown.

If Status: ELIGIBLE:
1) Detect headings and use markdown headings (#)
2) Split paragraphs
3) Remove unnecessary whitespace
4) Do not change content beyond that; do not add new content

OUTPUT FORMAT:

Status: [ELIGIBLE/NOT ELIGIBLE]
Reason: [Short explanation]

---MARKDOWN_START---
[ONLY if Status: ELIGIBLE, put the markdown-converted text here]
[If Status: NOT ELIGIBLE, leave this section COMPLETELY EMPTY]
---MARKDOWN_END---

URL: {url}
Text: {text}
\end{PromptListing}
\label{prompt:content_filter}
\end{promptbox}

\subsection{Topic Classification Prompt}
\begin{promptbox}{Topic Classification}
\begin{PromptListing}
You are labeling Turkish text for an LLM dataset. You must not use or infer any "source type" (Wikipedia vs web) in your decision. Treat every document the same.
You will be given one document: title (optional), url (optional), and text (may be truncated).
Task:

1) Assign exactly ONE topic category: topic_l1
2) Assign exactly ONE safety category: safety_label

Allowed values:
topic_l1 (choose exactly one):
- STEM
- Humanities
- Social_Sciences
- Professional_Applied
- Culture_Entertainment
- Everyday_Life
- Geography_Places
- Biography_People
- Organizations_Institutions
- Events_History
- Meta_Content

safety_label (choose exactly one):
- Safe
- Needs_Filtering
- Exclude
Safety guidelines:
- Safe: ordinary content with no clear policy risks.
- Needs_Filtering: contains potentially sensitive/age-restricted/controversial material or advisory content (e.g., medical or financial advice, explicit profanity/hate slurs contextually used, graphic descriptions) but not clearly disallowed.
- Exclude: clearly disallowed or high-risk content, such as explicit instructions for wrongdoing (e.g., making weapons, fraud), explicit sexual content involving minors, actionable self-harm instructions, doxxing/PII, extremist recruitment/praise, or pervasive hate/harassment.

Output rules:
- Output JSON only. No markdown. No extra keys.
- Keep rationale <= 200 characters, grounded only in the given text.

JSON format:

{
  "topic_l1": "...",
  "safety_label": "...",
  "rationale": "..."
}

Now label this document:
TITLE: {{title}}
URL: {{url}}

TEXT: {{text}}
\end{PromptListing}
\label{prompt:topic_classifier}
\end{promptbox}

\subsection{QA Generation Prompts}
\label{prompt:qa_generation}

We employ two types of prompts for question generation depending on the context: single-chunk and multi-chunk generation.

\subsubsection{Single Chunk Generation}
\begin{promptbox}{Single Chunk Generation}
\begin{PromptListing}
Generate exactly {num_questions} question-answer pair(s) that can be answered from this text chunk:

Chunk ID: {chunk_id}{context_info}

Text:
{chunk_content}

Each question must be categorized into one of these two categories:
1. **FACTUAL**: Questions that test direct recall of specific details. The answer is a specific name, date, number, or short verbatim phrase found directly in the text.
2. **INTERPRETATION**: Questions that test comprehension by asking for explanations of causes, effects, or relationships between concepts in the text. The answer requires synthesizing information rather than just quoting it.

Requirements:
- Only ask about information explicitly stated in this text
- Make questions specific and factual
- Each question should be answerable from this chunk alone
- Provide complete, accurate answers based solely on the chunk content
- Categorize each question appropriately based on the type of cognitive task required
- Return valid JSON with the specified structure
- Do NOT use markdown code blocks (like)
- Return ONLY the JSON object, no other text
\end{PromptListing}
\label{prompt:qa_single}
\end{promptbox}

\subsubsection{Multi-Chunk Generation}
\begin{promptbox}{Multi-Chunk Generation}
\begin{PromptListing}
Generate exactly {num_questions} question-answer pair(s) that require information from multiple chunks below.

These chunks are related. Generate questions that:
1. Require information from at least 2 of the provided chunks
2. Are about connections, relationships, comparisons, or broader concepts across chunks
3. Cannot be answered from any single chunk alone{context_info}

Each question must be categorized into one of these two categories:
1. **FACTUAL**: Questions that test direct recall of specific details. The answer is a specific name, date, number, or short verbatim phrase found directly in the text.
2. **INTERPRETATION**: Questions that test comprehension by asking for explanations of causes, effects, or relationships between concepts in the text. The answer requires synthesizing information rather than just quoting it.

Chunks:
{chunks_text}

Requirements:
- Focus on relationships and connections between the chunks
- Make questions that require synthesis of information
- Provide complete answers that synthesize information from multiple chunks
- Categorize each question appropriately based on the type of cognitive task required
- Return valid JSON with chunk IDs {chunk_ids} in related_chunk_ids
- Do NOT use markdown code blocks (like)
- Return ONLY the JSON object, no other text
\end{PromptListing}
\label{prompt:qa_multi}
\end{promptbox}

\subsection{QA Validation Rubric}
\begin{promptbox}{QA Validation System Prompt}
\begin{PromptListing}
You evaluate question-answer pairs for accuracy.

Check if:
- Question is clear
- Answer is accurate based on provided text chunks
- Answer fully addresses the question
- Chunks contain all necessary information

Return JSON: {"is_correct": boolean, "reason": "brief explanation"}

Keep reason concise (max 50 words). Return ONLY valid JSON.
\end{PromptListing}
\label{prompt:qa_validation}
\end{promptbox}

\section{Full Experimental Results and Algorithms}
\label{app:full_results}

\begin{algorithm}[h!] 
  \caption{Genetic search over modular RAG pipelines} \label{alg:ga_app} 
  \begin{algorithmic}[1] 
    \Require Families $\{\mathcal{M}_f\}_{f=1}^F$, population size $P$, generations $G$, mutation rate $\mu$, evaluation set $\mathcal{Q}$ 
    \State Initialize population $\mathcal{P}_0=\{g_i\}_{i=1}^P$ by sampling valid genomes 
    \For{$t=1$ to $G$} 
    \State Evaluate $\mathrm{Fit}(g)$ for all $g\in\mathcal{P}_{t-1}$ on $\mathcal{Q}$ 
    \State Select elites $\mathcal{E}$ and parents $\mathcal{S}$ (e.g., tournament selection) 
    \State Create offspring via crossover over genomes in $\mathcal{S}$ 
    \State Mutate genes with probability $\mu$
    \State Form $\mathcal{P}_t \leftarrow \mathcal{E} \cup \mathcal{O}$ 
    \EndFor 
    \State 
    \Return best genomes from $\mathcal{P}_G$ 
  \end{algorithmic} 
\end{algorithm}

\begin{table}[h]
\centering
\small
\caption{Top base domains by document frequency with cumulative coverage (Web Part).}
\label{tab:top_domains_app}
\begin{tabular}{lrr}
\toprule
\textbf{Domain} & \textbf{Docs} & \textbf{Cumulative \%} \\
\midrule
sikayetvar.com         & 227 & 3.6 \\
haberler.com           & 158 & 6.1 \\
posta.com.tr           & 134 & 8.2 \\
mynet.com              & 132 & 10.3 \\
donanimhaber.com       & 108 & 12.0 \\
webtekno.com           & 100 & 13.6 \\
onedio.com             & 98  & 15.2 \\
sondakika.com          & 98  & 16.7 \\
fanatik.com.tr         & 91  & 18.2 \\
haberaktuel.com        & 89  & 19.6 \\
\bottomrule
\end{tabular}
\end{table}

\begin{table*}[h!]
\centering
\caption{Complete performance results for \textbf{noteworthy} evaluated RAG configurations.}
\tiny
\setlength{\tabcolsep}{2pt}
\renewcommand{\arraystretch}{1.1}
\begin{tabularx}{\textwidth}{>{\raggedright\arraybackslash}X c c c c}
\toprule
\textbf{RAG Methods Combination} & \textbf{Overall Score} & \textbf{Retrieval} & \textbf{Generation} & \textbf{Total Token Usage} \\
\midrule
hyde + ce\_rerank + tree\_summarize + long\_context\_reorder & 85.00\% & 0.876 & 0.823 & 3,663.8 \\
hyde + ce\_rerank + llm\_summarize + reflection\_revising & 84.90\% & 0.876 & 0.822 & 3,118.4 \\
hyde + ce\_rerank + tree\_summarize + reflection\_revising & 84.80\% & 0.876 & 0.819 & 3,966.2 \\
ce\_rerank + adjacent\_augmenter + long\_context\_reorder & 84.60\% & 0.87 & 0.823 & 1,987.2 \\
hyde + tree\_summarize & 84.50\% & 0.892 & 0.798 & 5,260.4 \\
hyde + ce\_rerank + tree\_summarize + long\_context\_reorder + reflection\_revising & 84.50\% & 0.876 & 0.814 & 3,964.1 \\
hyde + ce\_rerank + adjacent\_augmenter + tree\_summarize + long\_context\_reorder & 84.40\% & 0.876 & 0.812 & 4,906.3 \\
ce\_rerank + adjacent\_augmenter + long\_context\_reorder & 84.40\% & 0.865 & 0.822 & 2,036.8 \\
hyde + tree\_summarize + long\_context\_reorder & 84.30\% & 0.892 & 0.794 & 5,276.3 \\
ce\_rerank + adjacent\_augmenter + llm\_summarize + long\_context\_reorder & 83.40\% & 0.87 & 0.799 & 2,703.2 \\
hyde + long\_context\_reorder + reflection\_revising & 83.10\% & 0.896 & 0.765 & 2,339.3 \\
hyde + llm\_rerank + tree\_summarize & 83.10\% & 0.868 & 0.795 & 4,295.2 \\
hyde + adjacent\_augmenter + long\_context\_reorder & 82.90\% & 0.896 & 0.761 & 3,138.8 \\
adjacent\_augmenter + long\_context\_reorder & 82.70\% & 0.896 & 0.758 & 2,147.0 \\
llm\_rerank + adjacent\_augmenter + llm\_summarize & 82.70\% & 0.863 & 0.792 & 2,973.8 \\
ce\_rerank + llm\_summarize + long\_context\_reorder & 82.40\% & 0.87 & 0.778 & 2,167.4 \\
ce\_rerank + long\_context\_reorder + reflection\_revising & 82.40\% & 0.865 & 0.783 & 1,773.4 \\
hyde + relevant\_segment\_extractor + llm\_summarize + long\_context\_reorder & 82.00\% & 0.891 & 0.75 & 2,685.5 \\
ce\_rerank + adjacent\_augmenter + tree\_summarize + long\_context\_reorder & 81.90\% & 0.865 & 0.772 & 4,762.7 \\
hyde + llm\_summarize + long\_context\_reorder & 81.50\% & 0.896 & 0.733 & 3,437.9 \\
simple\_query\_refinement\_clarification + ce\_rerank + adjacent\_augmenter + long\_context\_reorder & 81.10\% & 0.904 & 0.719 & 1,928.6 \\
adjacent\_augmenter + llm\_summarize + long\_context\_reorder & 81.10\% & 0.896 & 0.726 & 3,073.2 \\
hyde + llm\_summarize & 81.10\% & 0.896 & 0.726 & 3,409.4 \\
ce\_rerank + relevant\_segment\_extractor + llm\_summarize + long\_context\_reorder + reflection\_revising & 81.10\% & 0.87 & 0.753 & 2,828.2 \\
hyde + llm\_summarize & 80.90\% & 0.896 & 0.723 & 3,261.5 \\
hyde + adjacent\_augmenter + llm\_summarize + long\_context\_reorder & 80.70\% & 0.896 & 0.717 & 3,994.5 \\
hyde + llm\_summarize + long\_context\_reorder & 80.60\% & 0.896 & 0.715 & 3,472.5 \\
hyde + llm\_summarize + long\_context\_reorder + reflection\_revising & 80.50\% & 0.896 & 0.714 & 3,806.4 \\
ce\_rerank + relevant\_segment\_extractor + tree\_summarize + long\_context\_reorder & 80.50\% & 0.87 & 0.74 & 3,886.9 \\
simple\_query\_refinement\_clarification + ce\_rerank + adjacent\_augmenter & 80.20\% & 0.901 & 0.704 & 1,738.0 \\
query\_expansion\_simple\_multi\_query\_borda + ce\_rerank + adjacent\_augmenter & 80.00\% & 0.887 & 0.712 & 1,431.4 \\
llm\_rerank + llm\_summarize & 80.00\% & 0.885 & 0.715 & 2,707.5 \\
simple\_query\_refinement\_clarification + llm\_rerank + adjacent\_augmenter & 79.80\% & 0.903 & 0.693 & 2,291.5 \\
simple\_query\_refinement\_clarification + ce\_rerank + adjacent\_augmenter + reflection\_revising & 79.80\% & 0.9 & 0.697 & 1,991.6 \\
simple\_query\_refinement\_clarification + ce\_rerank + adjacent\_augmenter + tree\_summarize + long\_context\_reorder + reflection\_revising & 79.60\% & 0.836 & 0.756 & 4,524.5 \\
query\_expansion\_simple\_multi\_query\_borda + ce\_rerank + similarity\_threshold + adjacent\_augmenter + llm\_summarize + reflection\_revising & 79.40\% & 0.882 & 0.706 & 2,155.3 \\
hyde + ce\_rerank + llm\_summarize + long\_context\_reorder & 79.40\% & 0.877 & 0.711 & 3,197.8 \\
hyde + tree\_summarize + long\_context\_reorder & 79.30\% & 0.896 & 0.689 & 7,137.7 \\
hyde + ce\_rerank + adjacent\_augmenter + llm\_summarize & 79.30\% & 0.877 & 0.708 & 3,343.6 \\
vector\_simple + simple\_threshold + simple\_listing (Baseline) & 78.70\% & 0.872 & 0.702 & 1,000.4 \\
adjacent\_augmenter + tree\_summarize + long\_context\_reorder & 78.40\% & 0.896 & 0.672 & 7,216.6 \\
ce\_rerank + similarity\_threshold + relevant\_segment\_extractor + tree\_summarize + reflection\_revising & 78.30\% & 0.755 & 0.81 & 2,717.9 \\
\bottomrule
\end{tabularx}
\label{tab:full_results}
\end{table*}

\end{document}